\documentclass[12pt]{amsart}
\usepackage{amsbsy,amssymb,amsmath,amsthm,amscd,amsfonts,latexsym,amstext,delarray,
amsmath,graphicx} 
\usepackage[margin=.8in]{geometry}
\usepackage{color}

\numberwithin{equation}{section}

\def\F{{\mathbb F}}

\title[Prevalence and recoverability]{Prevalence and recoverability of syntactic parameters in sparse distributed memories}
\author[J.J.Park, R.Boettcher, A.Zhao, A.Mun, K.Yuh, V.Kumar, M.Marcolli]{Jeong Joon Park, 
Ronnel Boettcher, Andrew Zhao, Alex Mun, Kevin Yuh, \\ Vibhor Kumar, Matilde Marcolli}
\address{Division of Physics, Mathematics, and Astronomy \\ California Institute of Technology \\
1200 E. California Blvd, Pasadena, CA 91125, USA}
\email{jpark3@caltech.edu}
\email{ronnel@caltech.edu}
\email{azhao@dmail.caltech.edu}
\email{axlemn@gmail.com}
\email{kyuh@caltech.edu}
\email{vibhor@caltech.edu}
\email{matilde@caltech.edu}
\date{}

\begin{document}
\maketitle

\begin{abstract}
We propose a new method, based on Sparse Distributed Memory (Kanerva Networks), for studying dependency relations between different syntactic parameters in the Principles 
and Parameters model of Syntax. We store data of syntactic parameters
of world languages in a Kanerva Network and we check the recoverability of
corrupted parameter data from the network. We find that different syntactic
parameters have different degrees of recoverability. We identify two
different effects: an overall underlying relation between the prevalence of parameters across
languages and their degree of recoverability, and a finer effect that makes some parameters
more easily recoverable beyond what their prevalence would indicate. We interpret a higher 
recoverability for a syntactic parameter as an indication of the existence 
of a dependency relation, through which the given parameter can be 
determined using the remaining uncorrupted data.
\end{abstract}

\section{Introduction}

\medskip
\subsection{Syntactic Parameters of World Languages}\label{ParamSec}

The general idea behind the Principles and Parameters approach to Syntax, 
\cite{Chomsky}, \cite{ChoLa}, is the encoding of syntactic properties of natural
languages as a string of binary variables, the {\em syntactic parameters}. 
This model is sometimes regarded as controversial, and some schools of Linguistics 
have, consequently, moved towards other possible ways of modeling syntax. 
However, syntactic parameters remain more suitable than other
concurrent models from the point of view of a mathematical approach,
as we set out to demonstrate in a series of related papers \cite{Mar}, \cite{PGGCLDM},
\cite{STM}. Among the shortcomings ascribed to the Principles and Parameters model
(see for instance \cite{Hasp2}) 
is the fact that it has not been possible, so far, to identify a 
complete set of such syntactic parameters, even though
extensive lists of parameters are classified and recorded
for a large number of natural languages. It is also unclear what
relations exist between parameters and whether there is
a natural choice of a set of independent variables among them.

\smallskip

At present, sufficiently rich databases of syntactic parameters of world
languages are available, most notably the `Syntactic Structures of the World's 
Languages" (SSWL) database \cite{SSWL} (recently migrated to 
TerraLing \cite{Terraling}) and the ``World Atlas of Language 
Structures" (WALS) \cite{Hasp}. This makes it possible to reconsider the problem
of syntactic parameters, loosely formulated as understanding the geometry of
the parameter space and how parameters are distributed across language
families, with modern methods of data analysis. For example, topological data
analysis was applied to syntactic parameters in \cite{PGGCLDM}. In the present
paper, the main tool of analysis we will employ to study relations between
syntactic parameters will be Kanerva Networks. 

\smallskip

In this paper we selected a list of 21 syntactic parameters, mostly having to
do with word order relations (see \S \ref{listparamSec} below of a detailed
discussion of the chosen parameters), and a list of 166 languages, for which the
values of these parameters are recorded in the SSWL database (the languages
used are listed in the Appendix). The parameters are selected so that they
clearly are not an independent set of binary variables (see the discussion
in \S \ref{dependSec} below). The languages are selected so that they cut
across a broad range of different linguistic families. By storing the data of
syntactic parameters for this group of languages in a Kanerva Network, we
can test for recoverability when one of the binary variables is corrupted.
We find an overall relation between recoverability and prevalence across
languages, which depends on the functioning of the sparse distributed
memory. Moreover, we also see a further effect, which deviates from a
simple relation with the overall prevalence of a parameter. This shows
that certain syntactic parameters have a higher degree of recoverability in
a Kanerva Network. This property can be interpreted as a consequence of 
existing underlying dependence relations between different parameters.  With this
interpretation, one can envision a broader use of Kanerva Networks as
a method to identify further, and less clearly visible, dependence relations
between other groups of syntactic parameters.  

\smallskip

Another reason why it is interesting to analyze syntactic parameters using
Kanerva Networks is the widespread use of the latter as models of human
memory, \cite{Furb}, \cite{Kanerva2}, \cite{Keeler}. 
In view of the problem of understanding mechanism of language 
acquisition, and how the syntactic structure of language may be stored in
the human brain, sparse distributed memories appear to be a promising
candidate for the construction of effective computational models. 

\bigskip

\subsection*{Acknowledgment} This work was performed as part of the activities of the last author's 
Mathematical and Computational Linguistics lab and CS101/Ma191 class at Caltech. The last author 
is partially supported by NSF grants DMS-1201512 and PHY-1205440. 

\bigskip

\section{Syntactic Parameters}

\medskip
\subsection{Choice of parameters}\label{listparamSec}

For the purpose of this study, we focused on a list of 21 syntactic parameters,
which are listed in the SSWL database as
\begin{enumerate}
\item[01] Subject-Verb
\item[02] Verb-Subject
\item[03] Verb-Object
\item[04] Object-Verb
\item[05] Subject-Verb-Object
\item[06] Subject-Object-Verb
\item[07] Verb-Subject-Object
\item[08] Verb-Object-Subject
\item[09] Object-Subject-Verb
\item[10] Object-Verb-Subject
\item[11] Adposition-Noun-Phrase
\item[12] Noun-Phrase-Adposition
\item[13] Adjective-Noun
\item[14] Noun-Adjective
\item[15] Numeral-Noun
\item[16] Noun-Numeral
\item[17] Demonstrative-Noun
\item[18] Noun-Demonstrative
\item[19] Possessor-Noun
\item[20] Noun-Possessor
\item[A01] Attributive-Adjective-Agreement
\end{enumerate}

\medskip

The first $10$ parameters on this list deal with word order properties. 
{\em Subject-Verb} has the value~$1$ when in a clause with an intransitive 
verb the order subject followed by verb can be used in a neutral context, and value~$0$ 
otherwise. {\em Verb-Subject} has value~$1$ when, in the same setting,
the order verb followed by subject can be used. For example: English has value~$1$
for {\em Subject-Verb} and value~$0$ for {\em Verb-Subject} while Italian has value~$1$ 
for both parameters. {\em Verb-Object} has value~$1$ when a main verb (not the auxiliary) 
can precede its object in a neutral context, and $0$ otherwises; while {\em Object-Verb} has value~$1$
if the main verb can follow its object in a neutral context, and $0$ otherwise. 
English has {\em Verb-Object} value~$1$ and {\em Object-Verb} value~$0$;
German has value~$1$ for both; Japanese has {\em Verb-Object} set to $0$ and
{\em Object-Verb} value~$1$. The remaining $6$ parameters in this group describe
the different word order structures SVO, SOV, VSO, VOS, OSV, OVS: each of these
parameters has value~$1$ when the corresponding word order can be used in a
neutral context, and value~$0$ otherwise. These word order parameters have
very different distribution among the world languages: of the six possible word orders
listed above, it is estimated that around 45\% of the world languages follow the SOV order,
42\% the SVO, 9\% have VSO, 3\% have VOS, only 1\% follow the OVS order, and
the remaining possibility, OSV, is extremely rare, estimated at  only 0.2\%, see 
\cite{Tomlin}. We will return to discuss how the relative frequencies of different
parameters, within the group of languages that we consider in this paper, affect
the behavior in the Kanerva Network. 
The frequencies of the 21 parameters within the group of languages used for 
this study (see the list in the Appendix) are reported in the table below.

\begin{center}
\begin{tabular}{|c|c|}
\hline
Parameter & Frequency \\
\hline
\hline
[01] Subject--Verb  & 0.64957267 \\
\hline
[02] Verb--Subject & 0.31623933 \\
\hline
[03] Verb--Object & 0.61538464 \\
\hline
[04] Object--Verb & 0.32478634 \\
\hline
[05] Subject--Verb--Object & 0.56837606 \\
\hline
[06] Subject--Object--Verb & 0.30769232 \\
\hline
[07] Verb--Subject--Object & 0.1923077 \\
\hline
[08] Verb--Object--Subject & 0.15811966 \\
\hline
[09] Object--Subject--Verb & 0.12393162 \\
\hline
[10] Object--Verb--Subject & 0.10683761 \\
\hline
[11] Adposition--Noun--Phrase & 0.58974361 \\
\hline
[12] Noun--Phrase--Adposition & 0.2905983 \\
\hline
[13] Adjective--Noun & 0.41025642 \\
\hline
[14] Noun--Adjective & 0.52564102 \\
\hline
[15] Numeral--Noun & 0.48290598 \\
\hline
[16] Noun--Numeral & 0.38034189 \\
\hline
[17] Demonstrative--Noun & 0.47435898 \\
\hline
[18] Noun--Demonstrative & 0.38461539 \\
\hline
[19] Possessor--Noun & 0.38034189 \\
\hline
[20] Noun--Possessor & 0.49145299 \\
\hline
[A 01] Attributive--Adjective--Agreement & 0.46581197 \\
\hline
\end{tabular}
\end{center}

\smallskip

The {\em Adposition-Noun-Phrase} parameter is set to~$1$ in a language, when 
there are adpositions that precede the noun phrase they occurs with, while
the {\em Noun-Phrase-Adposition} parameter is set to~$1$ when there are adpositions
that follow  the noun phrase.
Both {\em Adposition-Noun-Phrase} and {\em Noun-Phrase-Adposition}
can have value~$1$ in a language that has both prepositions and
postpositions. The pair of parameters {\em Adjective-Noun} and {\em Noun-Adjective}
regulate whether an adjective can precede (respectively, follow) the noun it modifies 
in a neutral context. Similarly, {\em Numeral-Noun} and {\em Noun-Numeral} are set to~$1$
when there are, in the language, cardinal numerals that precede (respectively, follow) 
the noun they modify in a neutral context. The same for the pairs {\em Demonstrative-Noun}
and {\em Noun-Demonstrative}, and {\em Possessor-Noun} and {\em Noun-Possessor}
with respect to demonstratives (respectively, possessors) and the noun they modify.
Finally, the parameter {\em Attributive-Adjective-Agreement} is set to~$1$ for a language
when there are attributive adjectives that show agreement with (some of) the nouns 
they modify. For example, this parameter is $0$ for English and $1$ for Italian. 

\smallskip

A complete list of the syntactic parameters recorded in the SSWL database
and their linguistic meaning is available at
{\tt http://sswl.railsplayground.net/browse/properties}
and in TerraLing {\tt http://www.terraling.com/groups/9/properties}

\medskip

This particular choice of languages from the SSWL database is motivated by the fact
that, for this list, there is a complete mapping of the values of the 21 syntactic parameters
listed above. This makes it possible to construct a Kanerva network with enough data
points in it to carry out our intended analysis.

\medskip
\subsection{Parameters and Dependencies}\label{dependSec}

There is clearly some degree of dependence between the $6$ 
word order parameters SVO, SOV, VSO, VOS, OSV, OVS
and the previous $4$ parameters in the list, so that these cannot 
be all completely independent binary variables. However, this 
dependence relation is more subtle than it might appear at first. 
To illustrate the point with an example, consider the case of the 
languages English and Italian.
Both have $1$ for SVO and $0$ for VSO, but as mentioned above English has value~$1$
for {\em Subject-Verb} and value~$0$ for {\em Verb-Subject}, while Italian has value~$1$ 
for both parameters. This means that the relation between these parameters is not simply
a fixed algebraic dependence relation (unlike the entailment of parameters that we analyzed
in \cite{STM}, for example). Rather, there may be relations that are 
expressible probabilistically, in terms of frequencies and correlations. 
This is the type of relations that we seek to identify 
with the use of sparse distributed memories. 

\smallskip

Our purpose in this study is to determine how much the presence of dependencies between the
syntactic parameters is detectable through a Kanerva Network model, by measuring
recoverability of some parameters in terms of the remaining ones.

\medskip
\section{Sparse Distributed Memory}\label{SDMsec}

Kanerva Networks (or Sparse Distributed Memory) were developed by Pentti Kanerva in 1988, \cite{Kanerva},
\cite{Kanerva2}, as a mathematical model of human long term memory. The model allows for approximate
accuracy storage and recall of data at any point in a high dimensional space, using fixed hard locations 
distributed randomly throughout the space. During storage of a datum, hard locations ``close" to the datum
encode information about the data point. Retrieval of information at a location in the space is 
performed by pooling nearby hard locations and aggregating their encoded data. 
The mechanism allows for memory addressability of a large memory space with reasonable 
accuracy in a sparse representation.

\smallskip

Kanerva Networks model human memory in the following way: a human thought, perception, 
or experience is represented as an (input) feature vector -- a point in a high dimensional space. 
Concepts stored by the brain are also represented as feature vectors, and are usually stored 
relatively far from each other in the high dimensional space (the mind). 
Thus, addressing the location represented by the input vector will yield, to a reasonable 
degree of accuracy, the concept stored near that location. Thus, Kanerva Networks model 
the fault tolerance of the human mind -- the mind is capable of mapping imprecise input 
experiences to well defined concepts. For a short introduction to Kanerva Networks aimed
at a general public, see \S 13 of \cite{Fran}.

\smallskip

More precisely, the functioning of Kanerva Network models can be summarized as follows.
Over the field $\F_2=\{ 0,1 \}$, consider a vector space (Boolean space) $\F_2^N$ of sufficiently large
dimension $N$. Inside $\F_2^N$, choose a uniform random sample of $2^k$ hard locations, 
with $2^k << 2^N$. Compute the median Hamming distance between hard locations. The 
{\em access sphere} of a point in the space $\F_2^N$ is a Hamming sphere of radius 
slightly larger than this median value (see \S 6 of \cite{Kanerva} for some precise estimates).
When writing to the network at some location $X$ in the space $\F_2^N$, data is 
distributively stored by writing to all hard locations within the access sphere of that point $X$. 
Namely, each hard location stores $N$ counters (initialized to $0$), and all hard locations within 
the access sphere of $X$ have their $i$-­th counter incremented or decremented by $1$, 
depending on the value of the $i$-­th bit of $X$, see \S 3.3.1 of \cite{Kanerva2}. 
When the operation is performed for a set of
locations, each hard location stores a datum whose $i$-th entry is determined by the majority rule of 
the corresponding $i$-th  entries for all the stored data.
One reads at a location $Y$ in the network a new datum, whose $i$-­th entry is 
determined by comparing $0$ to the $i$-­th counters of all the hard 
locations that fall within the access sphere of $Y$, that is, the $i$-th entry 
read at $Y$ is itself given by the majority rule on the $i$-th entries of
all the data stored at all the hard locations accessible from $Y$.
For a more detailed account, see \cite{Kanerva}, \cite{Kanerva2}, and the 
summary in \S 13 of \cite{Fran}. 

\smallskip

The network is typically successful in reconstructing stored data, because intersections between
access spheres are infrequent and small. Thus, copies of corrupted data in hard locations within 
the access sphere of a stored datum $X$ are in the minority with respect
to hard locations faithful to $X$'s data. When a datum is corrupted by noise ({\em i.e.}~flipping 
bit values randomly), the network is sometimes capable of correctly reconstructing 
these corrupted bits. The ability to reconstruct certain bits hints that these bits are derived 
from the remaining, uncorrupted bits in the data.

\smallskip

In addition to modeling human memory in applications to neuroscience  and neural computation 
(see for instance \cite{Knob}), Kanerva networks have been used in various other contexts, 
such as weather prediction \cite{Rog}, robotics \cite{Men}, and as machine-­learning tools, 
in comparison to other forms of associative memory, \cite{Chou}, \cite{Hely}, \cite{Keeler}. Most 
applications of Kanerva networks in the literature have focused on models of memory 
and of data storage and recovery. While some applications to Linguistics
have been developed, for instance in the setting of speech recognition \cite{Prager}, Kanerva networks 
have not been previously used to analyze syntactic structures and identify dependencies
between syntactic parameters.

\medskip
\subsection{Detecting Parameter Dependencies}\label{DepSec}

Although Kanerva Networks were originally developed for and motivated by human memory, 
they are also a valuable general tool for detecting dependencies in a high-dimensional data sets. The reasons for this can be found in the literature on Kanerva Networks, see for 
instance the discussion in \cite{Hely}.

In the present paper, we treat each language, and its corresponding list of syntactic parameters, 
as a single data point in the network. Concretely, each data point is a concatenated binary 
string of all the values, for that particular language, 
of the 21 syntactic parameters  listed in \S \ref{ParamSec}.

As we recalled above, a Kanerva network operates by writing to uniformly random hard locations 
within a Hamming sphere of specified radius centered at the write location (specified by a bitstring), 
and reading from hard locations within a Hamming sphere centered at the read location, returning 
the majority rule derived from the data points for each of the individual bits. 

Regardless of how well this is representative of human memory, this system can demonstrate 
a clear correlation ({\em i.e.}~dependence) between certain parameters. Observe that, if we had 
written to clusters of data points in the space, interpreted as separate syntactic families of languages,
then reading from locations in the vicinity of the locations of these clusters would result in reading back 
a necessarily correlated set of parameter values, due to the each parameters being determined 
by the locally smaller set of hard locations. Here, by syntactic families, we do not necessarily mean
historical-linguistic families, but rather families of languages whose data set cluster together in
the Kanerva Network space. How well such groupings reflect historical-linguistic families remains
an issue for future investigation.
If the original location came from a cluster or family of languages, then we would expect to see 
corrupted bits recovered, indicating that this particular subset of bits is dependent on the rest, 
{\em i.e.}~that the parameters are not independent since there exists a non-­zero correlation 
between their values.

%%%
\begin{figure}
\begin{center}
\includegraphics[scale=0.72]{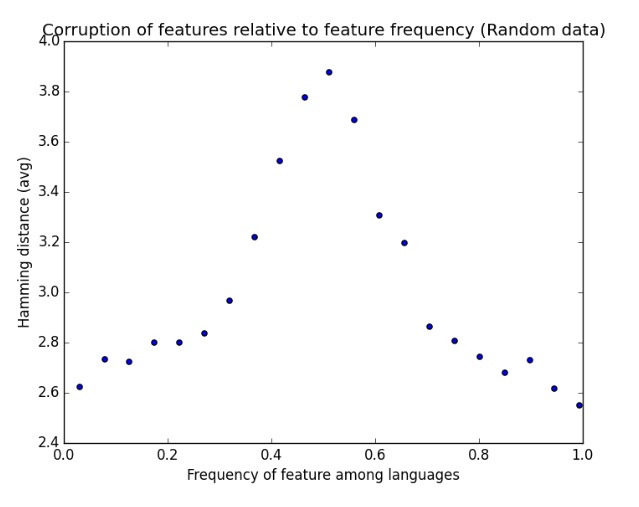}
\end{center}
\caption{Prevalence and recoverability in a Kanerva Network (random data).}
\label{randomFig}
\end{figure}
%%%

%%%
\begin{figure}
\begin{center}
\includegraphics[scale=0.9]{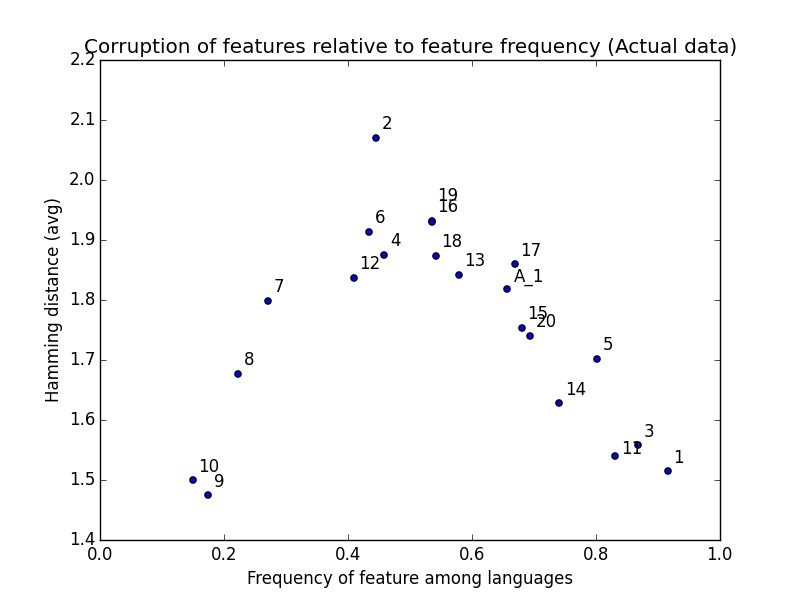}
\end{center}
\caption{Prevalence and recoverability for syntactic parameters in a Kanerva Network.}
\label{dataFig}
\end{figure}
%%%

\section{Implementation Method}\label{MethodSec}

We considered 166 languages from the SSWL database, which have a complete
mapping of the 21 syntactic parameters discussed in \S \ref{ParamSec}. These
provide 166 data points in a Kanerva Network with Boolean space $\F_2^{21}$.
The complete list of languages used is reported in the Appendix.

\smallskip

The {\tt python/c sdm} sparse distributed memory library\footnote{{\tt https://github.com/msbrogli/sdm}} 
was used to simulate the Kanerva network. The current state of the library at the time of the experiment 
was not functional, so the last working version from January 31, 2014 was used. The library was initialized 
with an access sphere of $n/4$, where $n$ is the median hamming distance between items. This was
the optimal value we could work with, because
larger values resulted in an excessive number of hard locations being in the sphere, which the library 
was unable to handle.

\smallskip

Three different methods of corruption were tested. First, the correct data was written to the 
Kanerva network, then reads at corrupted locations were tested. A known language bit-string, 
with a single corrupted bit, was used as the read location, and the result of the read was 
compared to the original bit-string in order to test bit recovery. The average Hamming distance 
resulting from the corruption of a given bit, corresponding to a particular syntactic
parameter, was calculated across all languages.

\smallskip

In order to test for relationships independent of the prevalence of the features, 
another test was run that normalized for this. For each feature, a subset of languages 
of fixed size was chosen randomly such that half of the languages had that feature. 
Features that had too few languages with or without the feature to reach the chosen 
fixed size were ignored for this purpose. For this test, a fixed size of 95 languages
was chosen, as smaller sizes would yield less significant results, and larger sizes 
would result in too many languages being skipped. The languages were then written 
to the Kanerva network and the recoverability of that feature was measured.

\smallskip

Finally, to check whether the different recovery rates we obtained for different syntactic
parameters were really a property of the language data, rather than of the 
Kanerva network itself, the test was run again with random data generated with an approximately 
similar distribution of bits. In this test, the general relationship of Figure \ref{randomFig} was observed.
This indicates that the general shape of the curve may be a property of the Kanerva network. 
The magnitude of the values for the actual data, however, is very different, see Figure \ref{dataFig}. 
This indicates that the recoverability rates observed for the syntactic parameters are begin
influenced by the language data, hence they should correspond to actual syntactic properties.

\smallskip

%A repository of the code used for this project is available at Bitbucket \newline
%\verb! https://bitbucket.org/insert_generic_username/cs101b_fnl !

\section{Summary of Main Results}\label{ResultSec}

Summarizing, the main results we obtained in the analysis of the
selected data of languages and parameters identifies two different
effects on the recoverability of syntactic parameters in Kanerva Networks.

\medskip
\subsection{Large scale structure: prevalence and recoverability}
The first effect is a general relation between prevalence of parameters
across languages and recoverability in sparse distributed memories.
This is a general effect that depends on the functioning of Kanerva
Networks and can be seen using random data with the same frequencies
as the chosen set of parameters. The curve expressing recoverability as
a function of prevalence using random data (Figure \ref{randomFig})
indicates the overall underlying effect. This phenomenon seems in
itself interesting, given ongoing investigations on how prevalence
rates of different syntactic parameters may correlate to neuroscience
models, see for instance \cite{Kemmerer}.

\medskip
\subsection{Smaller scale structures of recoverability}
In addition to the large scale relationship between prevalence of feature and recoverability
mentioned above, 
the variation of the recoverability values from the general trend is consistent and indicates 
a second order relationship, which we see in the plot of the real data of syntactic 
parameters in Figure~\ref{dataFig}. A far smaller variation from a smooth curve 
was observed when using random input data as in Figure \ref{randomFig}. 
The normalized test indicates a smaller but still significant variation 
in feature recoverability even when all features considered had the same 
prevalence among the dataset. 

\medskip
\subsection{Recoverability scores}
\begin{figure}
\begin{center}
\includegraphics[scale=0.72]{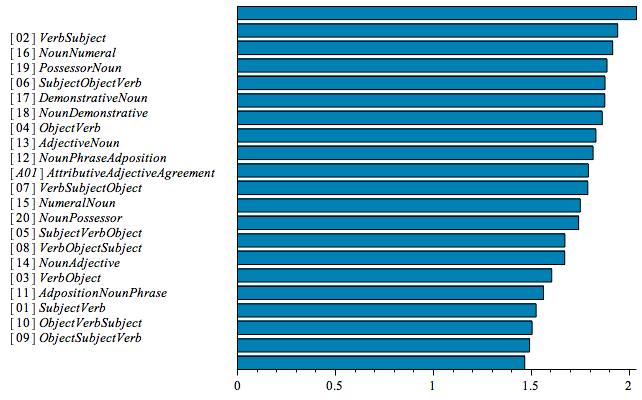}
\end{center}
\caption{Corruption of syntactic parameters in a sparse distributed memory (non-normalized). }
\label{BarChartFig}
\end{figure}

\begin{figure}
\begin{center}
\includegraphics[scale=0.72]{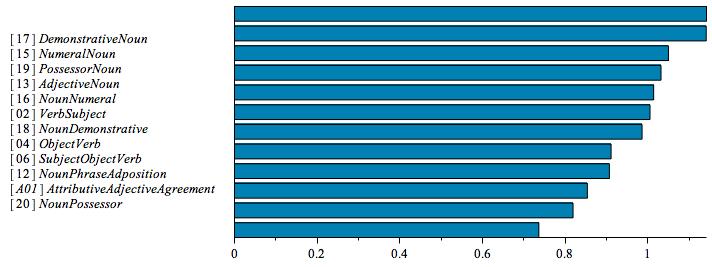}
\end{center}
\caption{Corruption (normalized test) of some syntactic parameters.}
\label{BarChartFigNorm}
\end{figure}

The resulting levels of recoverability of the syntactic parameters are listed in the table below,
and displayed in Figure~\ref{BarChartFig}. The results of the normalized test are listed, for
a selection of parameters, in the second table and displayed in 
Figure~\ref{BarChartFigNorm}.
To each parameter we assign a score, obtained by computing the average 
Hamming distance between the resulting bit-vector in the corruption experiment and the original one. The lower the
score, the more easily recoverable a parameter is from the uncorrupted data, hence from
the other parameters. 

\bigskip

\begin{center}
\begin{tabular}{|c|c|}
\hline
Parameter & Corruption (non-normalized) \\
\hline
\hline
[01] Subject-Verb & $1.50385541439$ \\ 
\hline
[02] Verb-Subject & $2.03638553143$ \\
\hline
[03] Verb-Object & $1.56180722713$ \\
\hline
[04] Object-Verb & $1.86186747789$ \\
\hline
[05] Subject-Verb-Object & $1.6709036088$  \\
\hline
[06] Subject-Object-Verb & $1.88596384645$ \\
\hline
[07] Verb-Subject-Object & $1.7879518199$ \\
\hline
[08] Verb-Object-Subject & $1.66993976116$ \\
\hline
[09] Object-Subject-Verb & $1.46596385241$ \\
\hline
[10] Object-Verb-Subject & $1.4907228899$ \\
\hline
[11] Adposition-Noun-Phrase & $1.52427710056$ \\
\hline
[12] Noun-Phrase-Adposition & $1.81512048125$ \\
\hline
[13] Adjective-Noun & $1.82927711248$ \\
\hline
[14] Noun-Adjective & $1.6037349391$ \\
\hline
[15] Numeral-Noun & $1.74969880581$ \\
\hline
[16] Noun-Numeral & $1.94036144018$ \\
\hline
[17] Demonstrative-Noun & $1.87596385121$ \\
\hline
[18] Noun-Demonstrative & $1.87463855147$ \\
\hline
[19] Possessor-Noun & $1.91487951279$ \\
\hline
[20] Noun-Possessor & $1.74102410674$ \\
\hline
[A01] Attributive-Adjective-Agreement &  $1.79102409244$\\
\hline
\end{tabular}
\end{center}

%%%
%%%

\begin{center}
\begin{tabular}{|c|c|}
\hline
Parameter & Corruption (normalized) \\
\hline
\hline
[02] Verb-Subject & $1.00494736842$ \\
\hline
[04] Object-Verb & $0.910842105263$ \\
\hline
[06] Subject-Object-Verb & $0.906736842105$ \\
\hline
[12] Noun-Phrase-Adposition & $0.853473684211$ \\
\hline
[13] Adjective-Noun & $1.03157894737$ \\
\hline
[15] Numeral-Noun & $1.14094736842$ \\
\hline
[16] Noun-Numeral & $1.01378947368$ \\
\hline
[17] Demonstrative-Noun & $1.14157894737$ \\
\hline
[18] Noun-Demonstrative & $0.985789473684$ \\
\hline
[19] Possessor-Noun & $1.04957894737$ \\
\hline
[20] Noun-Possessor & $0.736105263158$ \\
\hline
[A01] Attributive-Adjective-Agreement &  $0.818842105263$\\
\hline
\end{tabular}
\end{center}

%%%
%%%

\bigskip
\section{Further Questions and Directions}\label{FInalSec}

We outline here some possible directions in which we plan to expand the
present work on an approach to the study of syntactic parameters using
Kanerva Networks.

\subsection{Kanerva Networks and Language Families} 

Through our experiments of corrupting a syntactic parameter and checking whether the 
Kanerva Network can successfully reconstruct the original data, we have learned that the 
corruption of certain syntactic parameters is more fixable in the Kanerva Network. One 
interpretation of this result is that such parameters are dependent on the remaining ones. 
Indeed, for the set of syntactic parameters used in this study, we know a priori, for linguistic
reasons, that there should be a certain degree of dependency between some of 
the parameters, 
for example in the case of the first group of ten parameters governing the word order 
relations between subject, verb, and object, with the caveat discussed in \S \ref{dependSec}
above on how one should interpret such relations. A more detailed study of known  
relations between other groups of syntactic parameters and how they correlate to 
measures of recoverability in a Kanerva Network would be needed in order to 
better understand how syntactic dependencies affect recoverability, and further 
develop Kanerva Networks as a possible approach to detect additional dependency 
relations between the binary variables of other syntactic parameters. 

\smallskip

As we have seen,  the scalar score we obtain from the corruption experiments 
indicates how tractable is a variable,  or syntactic parameter, in the context of 
data points in its vicinity. In other words, if the scalar 
score is small for a certain parameter, then the parameter is derivable from other correct bits. 
Yet, one limitation of our result is that this scalar score is simply computed as 
the average of the Hamming distance between the resultant bit-vector and the 
original bit-vector. The derivability of 
a certain parameter might vary depending on the family of languages that it belongs to. 
For example, when a certain language feature is not robust to corruption in certain regions 
of the Kanerva Network, which means the parameter is not depended on other parameters, 
but robust to corruption in all the other regions, we will get a low scalar score. 

\smallskip

While our present approach can provide some meaningful insight about whether a 
certain feature is generally retrievable by analyzing other features, it does not shed 
light on identifying which feature is a determining feature in a family of languages. 
In other words, if a feature is very tractable (low scalar score) in one family of languages, 
this means that feature is a sharing characteristic of the language group. If it is not very 
tractable, then it might indicate that the feature is a changeable one in the group. 
Thus, by conducting the same experiments grouped by language families, we may
be able to get some information about which features are important in which language family.

\smallskip

It is reasonable to
assume that languages belonging to the same historical-linguistic family are located near each 
other in the Kanerva Network. However, a more detailed study where data are broken down
by different linguistic families will be needed to confirm this hypothesis. 

\smallskip

Under the assumption that closely related languages remain near in the Kanerva Network, 
the average of dependencies of a given parameter over the whole space might 
be less informative globally, 
because there is no guarantee that the dependencies would hold throughout 
all regions of the Kanerva Network. 
However, this technique may help identifying specific relations between 
syntactic parameters that hold within specific language families, rather than universally across all languages. The existence of such
relations is consistent with the topological features identified in \cite{PGGCLDM} which vary
across language families, so we expect to encounter similar phenomena from the Kanerva
Networks viewpoint as well. 

\medskip
\subsection{Kanerva Networks and the Language--Neuroscience Connection}\label{NeuroSec}

One of the main open frontiers in understanding human language is 
relating the structure of natural languages to the neuroscience of the
human brain. In an idealized vision, one could imagine a Universal
Grammar being hard wired in the human brain, with syntactic
parameters being set during the process of language acquisition (see \cite{Baker}
for an expository account).
This view is often referred to as the Chomskian paradigm, because it
is inspired by some of Chomsky's original proposals about Universal Grammar.
There have been recent objections to the Universal Grammar model,
see for instance \cite{Everett}. Moreover, a serious difficulty lies in the fact 
that there is, at present, no compelling evidence from the 
neuroscience perspective that would confirm this elegant idea. Some
advances in the direction of linking a Universal Grammar model of human
language to neurobiological data have been obtained in recent years: for
example, some studies have suggested Broca's area as a biological substrate for 
Universal Grammar, \cite{Marcus}.

\smallskip

Moreover, recent studies like \cite{Kemmerer} have found
indication of a possible link between the cross linguistic prevalence of 
syntactic parameters relating to word order structure and neuroscience
models of how action is represented in Broca's area of the human brain.
This type of results seems to cast a more positive light on the possibility
of relating syntactic parameters to computational neuroscience models.

\smallskip

Models of language acquisition based on neural networks have been
previously developed, see for example the survey \cite{PoVe}. Various
results, \cite{Chou}, \cite{Hely}, \cite{Kanerva3}, \cite{Keeler}, \cite{Knob},
have shown advantages of Kanerva's sparse distributed memories over
other models of memory based on neural networks. To our
knowledge, Kanerva Networks have not yet been systematically used
in models of language acquisition, although the use of Kanerva
Networks is considered in the work \cite{MacWhinney} on emergence 
of language. Thus, a possible way to extend the present model will be storing data
of syntactic parameters in Kanerva Network, with locations
representing (instead of different world languages) events in 
a language acquisition process that contain parameter-setting cues. 
In this way, one can try to create a model of parameter setting in language
acquisition, based on sparse distributed memories as a 
model of human memory. We will return to this approach in future work.

\bigskip

\section*{Appendix: Languages}

The list of languages from the SSWL database that we considered for
this study consists of: Acehnese, Afrikaans, Albanian,
American Sign Language, Amharic, Ancient Greek, Arabic (Gulf), 
Armenian (Eastern), Armenian (Western), Bafut, Bajau (West Coast), Bambara, 
Bandial, Basaa, Bellinzonese, Beng, Bengali, Bole, Brazilian Portuguese, 
Breton, Bulgarian, Burmese, Calabrian (Northern), Catalan, Chichewa,
Chol, Cypriot Greek, Czech, Dagaare, Digo, Digor Ossetic, 
Dutch, Eastern Armenian, English, English (Singapore), European Portuguese, Ewe,
Farefari, Faroese, Finnish, French, Frisian (West Frisian), Ga, Galician, Garifuna, Georgian,
German, Ghomala', Greek, Greek (Cappadocian), Greek (Homeric), 
Greek (Medieval), Gungbe (Porto­-Novo), Gurene, Gu\'ebie, Haitian,
Hanga, Hausa, Hebrew, Hindi, 'Hoan, Hungarian, Ibibio, Icelandic, Iha, Ilokano,  Imbabura Quichua
Indonesian, Irish, Iron Ossetic, Italian, Italian (Ancient Neapolitan), 
Japanese, K'iche', Karachay, Kashaya, Kayan, Khasi, 
KiLega, Kinande, Kiswahili, Kiyaka, Kom, Korean, Kuot,
Kurdish (Sorani), Kusunda, Lango, Lani, Lao, 
Latin, Latin (Late), Lebanese Arabic, Lubukusu, Maasai (Kisongo), 
Malagasy, Mandarin, Maori, Marshallese, Masarak, 
Medumba, Middle Dutch, Miya, Moroccan Arabic,
Muyang, Nahuatl (Central Huasteca), Naki, 
Nawdm, Ndut, Nepali, Northern Thai, Norwegian,
Nupe, Nweh, Okinawan, Old English, Old French,
Old Saxon, Oluwanga, One, Palue,  Panjabi,
Papuan Malay, Pashto, Pima, Polish, %Portuguese, 
Q'anjob'al, Romanian, Russian, Salasaca Quichua, Samoan,
San Dionisio Ocotepec Zapotec, Sandawe, 
Saweru, Scottish Gaelic, Senaya, Shupamem, 
Sicilian, Skou, Slovenian, Spanish, Swedish, Tagalog,
Taiwanese Southern Min, Thai, Tigre, Titan, Tlingit, Tommo­-So, Tongan
Triqui Copala, Tukang Besi, Tuki (Tukombo), Tupi (Ancient), Turkish,
Twi, Ukrainian, Vata, West Flemish, Wolane, Wolof, 
Yawa, Yiddish, Yoruba, Zulu.

\end{document}